\pdfoutput=1

\documentclass[11pt]{article}

\usepackage[preprint]{acl}

\usepackage{times}
\usepackage{latexsym}
\usepackage{microtype}
\usepackage{graphicx}
\usepackage{caption}
\usepackage{subcaption}
\usepackage{booktabs} 
\usepackage{tabularx}
\usepackage{amsmath}
\usepackage{amssymb}
\usepackage{mathtools}
\usepackage{amsthm}
\usepackage{lipsum}  
\usepackage{listings}
\usepackage{CJKutf8}
\usepackage{xspace}
\usepackage{xcolor}
\usepackage{multirow}
\usepackage[framemethod=TikZ]{mdframed}
\usepackage{tabu}
\usepackage{longtable}
\usepackage{alphalph}
\usepackage{bm}
\usepackage[subtle]{savetrees}

\newcolumntype{C}{>{\centering\arraybackslash}X}

\definecolor{ForestGreen}{rgb}{0.1333,0.545,0.1333}
\definecolor{Firebrick}{rgb}{0.698,0.1333,0.1333}

\usepackage{caption}
\usepackage{placeins}

\definecolor{ForestGreen}{rgb}{0.1333,0.545,0.1333}
\definecolor{Firebrick}{rgb}{0.698,0.1333,0.1333}

\usepackage{xcolor}  

\definecolor{robotaction}{RGB}{255, 140, 0}  
\definecolor{robottype}{RGB}{178, 34, 34}
\definecolor{robottask}{RGB}{65, 105, 225}  
\definecolor{controltype}{RGB}{0. 205, 0}
\definecolor{predsteps}{RGB}{216, 191, 216}
\definecolor{lavendermist}{rgb}{0.9, 0.9, 0.98}
\definecolor{visualtrace}{RGB}{255, 216, 0} 
\usepackage{colortbl}
\usepackage{xcolor}
\definecolor{lightgray}{gray}{0.9}
\usepackage{booktabs}
\usepackage{color, soul}
\usepackage{graphicx}
\usepackage{amsmath}
\usepackage{empheq}

\definecolor{lightgray}{gray}{0.9}
\definecolor{lightblue}{rgb}{0.93,0.95,1.0}
\definecolor{darkgreen}{rgb}{0.0,0.6,0.0}
\definecolor{darkblue}{rgb}{0.0,0.0,0.5}
\definecolor{pinegreen}{rgb}{0.0, 0.47, 0.44}
\definecolor{deepmagenta}{rgb}{0.8, 0.0, 0.8}
\definecolor{amber}{rgb}{1.0, 0.49, 0.0}

\newcommand{\ignorebig}[1]{}

\newcommand{\minisection}[1]{\noindent{\textbf{#1}.}}
\newcommand{\tabref}[1]{Table~\ref{#1}}

\newcommand{\figref}[1]{Figure~\ref{#1}}

\newlength\savewidth

\newcommand{\model}{Instruct-Verify-and-Act}
\newcommand{\smodel}{IVA}

\definecolor{citecolor}{RGB}{34,139,34}
\definecolor{lightred}{RGB}{241,140,142}
\definecolor{amber(sae/ece)}{rgb}{1.0, 0.49, 0.0}
\definecolor{battleshipgrey}{rgb}{0.52, 0.52, 0.51}
\definecolor{cadmiumorange}{rgb}{0.93, 0.53, 0.18}
\definecolor{applegreen}{rgb}{0.55, 0.71, 0.0}
\definecolor{cadmiumgreen}{rgb}{0.0, 0.42, 0.24}
\definecolor{forestgreen}{rgb}{0.13, 0.55, 0.13}
\definecolor{red}{rgb}{0.89, 0.0, 0.13}

\usepackage[T1]{fontenc}

\usepackage[utf8]{inputenc}

\usepackage{microtype}

\usepackage{inconsolata}

\usepackage{graphicx}

\title{Do What? Teaching Vision-Language-Action Models to Reject the Impossible}

\author{
  Wen-Han Hsieh\thanks{\ \ Equal contribution} \quad 
  Elvis Hsieh\footnotemark[1] \quad 
  Dantong Niu \quad 
  Trevor Darrell \quad 
  Roei Herzig \quad 
  David M. Chan \\
  University of California, Berkeley \\
  \texttt{\{hense1219, htelvis92, niudantong.88, trevordarrell, roeiherz, davidchan\}@berkeley.edu}
}

\begin{document}
\maketitle
\begin{abstract}

Recently, Vision-Language-Action (VLA) models have demonstrated strong performance on a range of robotic tasks. These models rely on multimodal inputs, with language instructions playing a crucial role-not only in predicting actions, but also in robustly interpreting user intent, even when the requests are impossible to fulfill. 
In this work, we investigate how VLAs can recognize, interpret, and respond to false-premise instructions—natural language commands that reference objects or conditions absent from the environment.
We propose ---{\model} ({\smodel}) --- a unified framework that (i) detects when an instruction cannot be executed due to a false premise, (ii) engages in language-based clarification or correction, and (iii) grounds plausible alternatives in perception and action. 
Towards this end, we construct a large-scale instruction tuning setup with structured language prompts and train a VLA model capable of handling both accurate and erroneous requests. Our approach leverages a contextually augmented, semi-synthetic dataset containing paired positive and false-premise instructions, enabling robust detection and natural language correction.
Our experiments show that {\smodel} can improves false premise detection accuracy by 97.56\% over baselines, while increasing successful responses in false-premise scenarios by 50.78\%.

\end{abstract}

\section{Introduction}

\begin{figure}
    \centering
    \includegraphics[width=\linewidth]{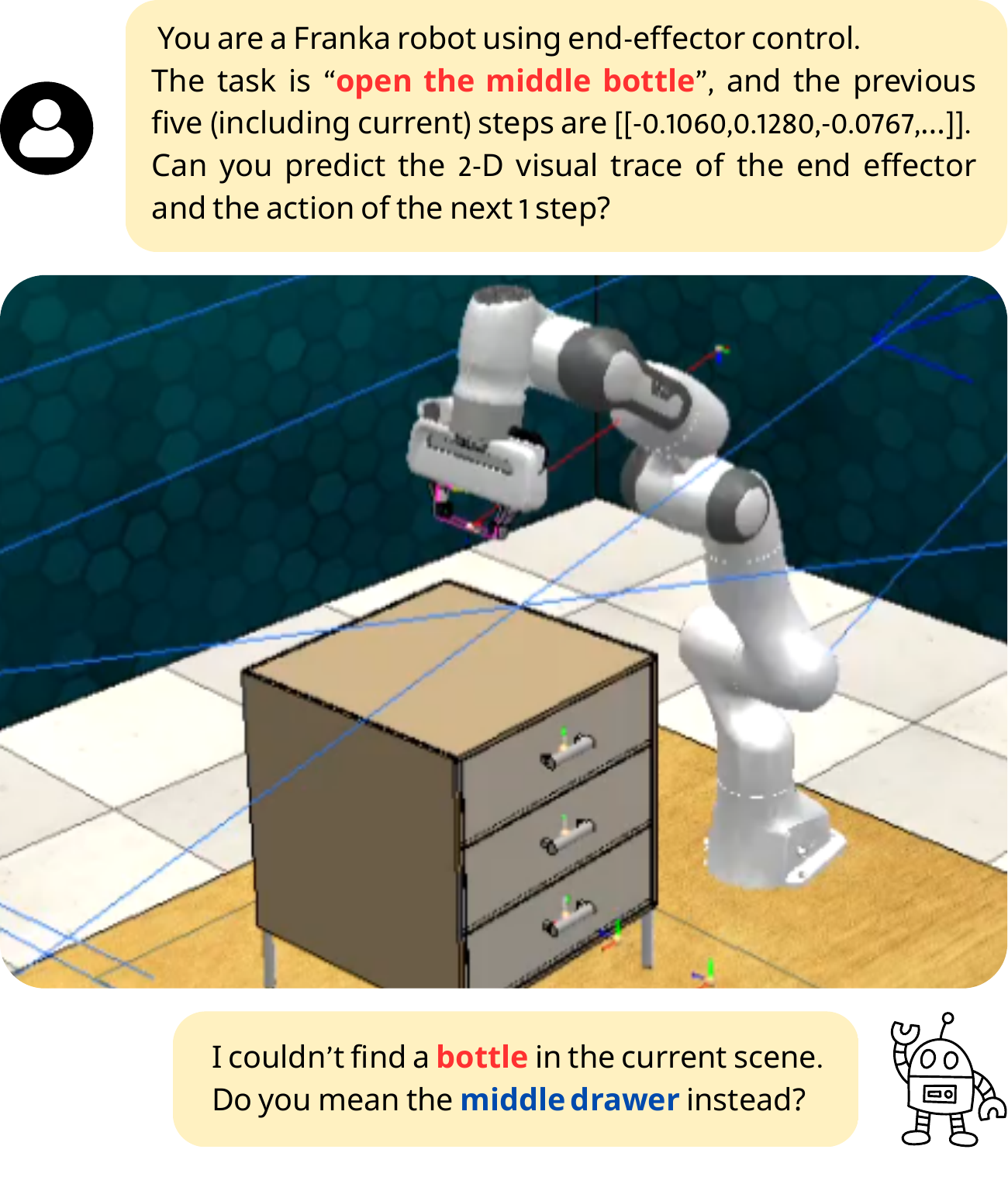}
    \caption{\textbf{Our {\model} ({\smodel}) framework is designed to handle false-premise instructions.} When the robot receives a command referencing a non-existent object (a \textit{\textcolor{red}{bottle}}), it detects the false premise, and generates a clarifying response that corrects the instruction, suggesting a valid alternative (a \textit{\textcolor{darkblue}{drawer}}).}
    \label{fig:teaser}
\end{figure}

Vision-Language-Action (VLA) models~\cite{Niu2024, kimOpenVLAOpenSourceVisionLanguageAction2024, liLLaRASuperchargingRobot2024,niu2025arm4r} represent a significant advancement in robotics, enabling agents to perform tasks using multimodal inputs by integrating visual perception, natural language understanding, and action generation. However, as these models are increasingly deployed in open-ended, real-world environments, they must handle diverse and often ambiguous instructions issued by users. Consider a household robot presented with the command, ``Bring me the red mug on the kitchen table,'' when no such mug exists. In this scenario, the robot’s ability to recognize the impossibility of the request, explain the issue, and suggest plausible alternatives is essential for safe and effective human-robot interaction. Yet, most existing VLAs lack mechanisms to detect or respond appropriately to instructions grounded in false premises - commands that reference objects, attributes, or relations not present in the environment.


While large multimodal models have made significant strides in visual grounding and instruction following, prior research in both natural language processing and robotics has typically assumed that user instructions are feasible and contextually grounded. In domains such as visual question answering and dialogue, models have been studied for their robustness to unanswerable or contradictory queries have been developed which can evaluate and correct for false premises in the case of question answering \cite{johnson2017clevr, suhr2017corpus, liu2019clevr, hudson2019gqa, Gurari2018, mahendru2017promise,whitehead2022reliable, mashrur2023robust, mashrur2022semantic, li2020neural, prabhakar2018question, mahendru2017promise, Karnik2024EmbodiedRT, knowno}. However, in the context of embodied agents and robotics, the issue of handling false-premise instructions - recognizing unfulfillable commands and producing helpful language-based corrections - remains largely unexplored. Existing robotics benchmarks and VLAs focus on execution success given correct instructions, without testing how these systems respond when user intent is unattainable or ambiguous.

To address this gap, we develop a unified VLA  model capable of interpreting and correcting false-premise instructions in robotic settings. Our framework combines large-scale instruction tuning with contextually-augmented datasets, enabling the model to detect unfulfillable requests, provide natural language feedback, and suggest alternative courses of action. See~\figref{fig:teaser}. We evaluate our approach across eight robotic tasks, measuring false premise detection accuracy and the rate of successful responses in false-premise scenarios. Our method achieves an improvement of 97.56\% in false premise detection over established baselines, and increases success in false-premise handling by 50.78\%. These results demonstrate that language-aware robots can move beyond simple execution - reasoning about user intent, clarifying ambiguity, and interacting naturally even when confronted with the impossible.

\section{Related Work}

\minisection{Vision-Language-Action Models (VLAs)} VLAs integrate visual perception, language understanding, and action generation to produce robot control sequences from visual observations and language instructions. Models such as LLARVA~\cite{Niu2024}, $\pi$ \cite{intelligence2025pi}, Gemini Robotics \cite{team2025gemini}, OpenVLA~\cite{kimOpenVLAOpenSourceVisionLanguageAction2024}, RT-2~\cite{Brohan2023}, PaLM-E embodied multimodal model~\cite{Driess2023}, and LLaRA~\cite{liLLaRASuperchargingRobot2024} fine-tune pretrained large multimodal models (LMMs) to predict actions, often using special tokens to represent the action space. 
These models share the goal of adapting LMMs for robotic control as they differ in the choice of LMM and action encoding methods. While existing VLAs typically leverage language decoders pretrained for high-level tasks (e.g., image captioning~\cite{kimOpenVLAOpenSourceVisionLanguageAction2024} and VQA~\cite{brohanRT2VisionLanguageActionModels2023}), when deployed in diverse, real-world environments, they exhibit a critical limitation: the inability to handle instructions grounded in false premises. Our work addresses this gap by explicitly training VLA models to recognize when referenced objects or conditions are absent and to respond appropriately.





\minisection{False Premises} The problem of detecting false premises has long been studied in NLP, particularly in QA benchmarks such as SQUAD 2.0 \cite{Rajpurkar2018} and False QA \cite{Hu2023}, with research primarily focused on model uncertainty for unanswerable questions \cite{Raina2022, Sulem2022}. This concept of false premises has recently spread from NLP to other areas of interest, including visual question answering \cite{ray2016question}, image/text matching \cite{feng2012automatic,xu2015show,ordonez2011im2text,karpathy2015deep,fang2015captions}, image-grounded conversation \cite{mostafazadeh2017image}, tool usage \cite{toor2019question} and hallucination detection \cite{rohrbach2018object}. Research in human-robot interaction has explored handling imperfect instructions and clarifying ambiguous instructions \cite{Deits2013,Park2024,prabhakar2018question,Shi2024,Tellex2014}, however this short paper is the first to explore false premise tasks in the context of vision-language-action models (VLAs).

\section{A VLA that Instructs, Verifies, and Acts}
\label{subsec:Preliminaries}
\label{subsec:training}

Our method builds upon the LLARVA model~\cite{Niu2024}, a VLA architecture designed for robotic instruction following. LLARVA leverages instruction-tuned large multimodal models (LMMs) to jointly interpret visual observations, natural language instructions, and robot proprioception, outputting robot actions along with intermediate visual representations called \textit{visual traces}.

\minisection{Input} 
LLARVA takes a visual observation \( o_t \), an RGB image at timestep \( t \), and a structured natural-language instruction \( l_t \) as input. The language instruction encodes the robot type (e.g., Franka Panda), control mode (e.g., end-effector or joint control), task description (e.g., ``close the drawer''), proprioceptive states (internally-sensed joint-angle vectors that indicate the robot's current position before it plans the next move) from the previous \( h \) timesteps, and the number of future actions (\( n \)) to predict, the instruction tuning template is shown as follows:
\begin{quote}
\small
\texttt{``You are a [Robot] robot using [Control Mode] control. The task is [Task Description], and the previous [h] steps are [Proprioceptive States]. Can you predict the trajectory of the end-effector and the action of the next [n] steps?''}
\end{quote}

\paragraph{Model.} LLARVA integrates three main components as follows: 
\begin{itemize}
    \item \textbf{Vision Encoder:} A frozen pretrained visual encoder (CLIP ViT-L/14) encodes image observations \( o_t \) into visual tokens.
    \item \textbf{Language Encoder:} Tokenizes and embeds the language instruction \( l_t \), forming language tokens.
    \item \textbf{Multimodal Decoder:} An autoregressive transformer decoder combines visual and language tokens, generating predictions for robotic actions \( A_{t:t+n-1} \) and future visual traces \( P_{t:N} \), formally:
\end{itemize}
\[
\pi(o_t, l_t) \rightarrow A_{t:t+n-1}, P_{t:N}
\]

where \( A_{t:t+n-1} \) represents predicted robot actions for the next \( n \) steps, and \( P_{t:N} \) indicates predicted 2-D visual trajectories of the robot's end-effector from timestep \( t \) to episode end \( N \).

LLARVA is initially pretrained on large-scale vision-action instruction data from Open X-Embodiment (OXE) and subsequently fine-tuned for specific robotic tasks, enabling strong generalization across diverse tasks and environments.

\minisection{False Premise Instruction Dataset} Following \citet{Niu2024}, we utilize image-action pairs from the OXE dataset~\cite{vuong2023open}, providing rich visual, language, and action representations. However, prior work did not explicitly address reasoning about the feasibility of user instructions. To address this gap, we constructed a dedicated dataset explicitly curated to handle false premise scenarios and their corresponding corrections.

Our false-premise instruction dataset is generated from robotic trajectories available in RLBench \cite{james2019rlbenchrobotlearningbenchmark}. Each task in our dataset includes two categories of false premise instructions:

\begin{table*}
\small
\centering
\caption{Comparison of IVA and LLaRVA across RLBench tasks: overall success rate, FP (false premise) detection rate (In-Domain/Out-of-Domain), and TP (true premise) success rate. The overall success rate is computed as the average of true and false premises success rates.} 
\begin{tabularx}{\textwidth}{l
  *{3}{>{\centering\arraybackslash}p{1.8cm}}
  *{3}{>{\centering\arraybackslash}p{1.8cm}}
}
\toprule
\multirow{2}{*}{\textbf{Task}} & 
\multicolumn{3}{c}{\textbf{IVA}} & 
\multicolumn{3}{c}{\textbf{LLaRVA}} \\
\cmidrule(lr){2-4} \cmidrule(lr){5-7}
& Overall Success & FP Detection (In-Domain/Out-of-Domain) & TP Success & Overall Success & FP Detection (In-Domain/Out-of-Domain) & TP Success \\
\midrule
meat off grill    & 58\% & 100\% / 100\% & 16\% & 2\% & 0\% / 0\% & 4\% \\
open drawer       & 61\% & 100\% / 80\% & 32\% & 20\%  & 0\% / 0\% & 40\% \\
push buttons      & 68\% & 100\% / 100\%  & 36\% & 16\%   & 0\% / 0\%        & 32\%   \\
put money in safe & 64\% & 100\% / 100\%           & 28\%   & 20\%   & 0\% / 0\%        & 40\%   \\
reach and drag    & 80\% & 100\% / 100\%           & 60\%   & 22\%   & 0\% / 0\%        & 44\%   \\
slide block       & 96\% & 100\% / 100\% & 92\% & 44\%   & 0\% / 0\%        & 88\%   \\
sweep to dustpan  & 94\% & 100\% / 100\%           & 88\%   & 30\%   & 0\% / 0\%        & 60\%   \\
turn tap          & 61\% & 100\% / 80\%           & 32\%   & 20\%   & 0\% / 0\%        & 40\%   \\
close jar         & 50\%      & 100\% / 100\% & 0\% & 0\%   & 0\% / 0\%        & 0\%   \\
\bottomrule
\label{tab:results}
\end{tabularx}
\end{table*}

\minisection{In-Domain False Premise} These instructions involve geometrically similar and contextually plausible objects derived from related tasks, making the intended correction relatively intuitive. For instance, in the task of closing a jar, if the user prompt is ``The task is \textit{close the blue safe}'', the model is expected to respond, ``I don't see a safe in the current scene. Do you mean jar?''.

\minisection{Out-of-Domain False Premise} These instructions contain clearly infeasible or nonsensical requests involving objects or scenarios impossible within the given context. For example, during the open drawer task, if the user prompts "The task is \textit{open the top elephant}," the model should identify the absurdity and respond appropriately, such as "I couldn't find a elephant in the current scene," subsequently terminating the interaction since the request is fundamentally invalid.

For training purposes, our dataset composition strategically includes approximately 20\% of episodes containing Out-of-Domain false premises, and 65\% of episodes containing In-Domain false premises injected into 10\% of their respective steps. This deliberate distribution ensures comprehensive exposure and training of the model in recognizing and handling various types of erroneous instructions.

\minisection{{\smodel} Training} We adopt an end-to-end instruction-tuning approach, closely following LLARVA's training methodology, to train our model using the newly curated false-premise dataset. While keeping both the vision and language encoders frozen, we fine-tune the auto-regressive transformer decoder using standard LoRA adapters. 

Specifically, our training procedure utilizes 800 episodes per task, with each episode containing image observations \(o_t\), language instructions \(l_t\), ground-truth robotic actions \(\hat{A}_{t:t+n-1}\), and visual traces \(\hat{P}_{t:N}\). The training data includes a mixture of true-premise and false-premise instructions, with roughly 20\% of episodes containing Out-of-Domain false premises and approximately 65\% containing In-Domain false premises, introduced at 10\% of steps within each episode. Given \(o_t\) and \(l_t\), the model predicts actions and visual traces auto-regressively: 
$$
p(\hat{A}_{t:t+n-1}, \hat{P}_{t:N}\mid o_t, l_t) = \prod_{i=1}^{|R|} p_\theta(x_i \mid o_t, l_t)
$$

where \(\theta\) denotes trainable parameters, \(x_i\) is the predicted token at timestep \(i\), and \(R\) represents the full response sequence. We compute the training loss as the standard cross-entropy between predicted tokens and ground-truth annotations.

Unlike LLARVA's two-stage procedure (pre-training and fine-tuning), we train our model end-to-end on our unified dataset, ensuring simultaneous learning of accurate robotic action prediction, robust false-premise detection, and appropriate language-based correction responses.

\section{Results}
We evaluated {\smodel} on two aspects: its ability to detect and correct false-premise (FP) instructions, and its performance on standard, true-premise (TP) tasks.

\paragraph{Experiment Setup.} We conducted experiments on 9 RLBench tasks, generating 25 episodes per task with randomly varied object positions. Each episode was paired with two type of user prompts: one standard and one containing a false premise. During fine-tuning, the model received the front camera view and the previous 5 joint positions as input,
and predicted both the visual trace and the next action step, represented as an 8-dimensional vector (7 joint velocities plus a binary gripper state). We report the success/failure rates for one fixed seed per task on the validation set, comparing with LLARVA as a baseline. The results are shown in~\tabref{tab:results}.

\minisection{Evaluation Procedure}
We evaluate {\smodel} in a single‐pass, end‐to‐end setting on all 225 episodes (25 episodes × 9 RLBench tasks) with randomized object poses. For each episode, we generate a true‐ or false‐premise instruction and score IVA’s full response as follows:

\noindent (1) \textbf{Detection Stage.}  
IVA first outputs a textual response, which we parse to classify the instruction as either \emph{Accept} (True-Premise) or \emph{Clarify/Refuse} (False-Premise).  
  \begin{itemize}
    \item \textbf{Accept (True-Premise):} 
      Scored 1 if the subsequent executed trajectory exactly matches the ground-truth; otherwise 0.
    \item \textbf{Clarify/Refuse (False-Premise):} 
    Scored 1 for an explicit out-of-domain refusal (e.g., “I’m sorry, that object isn’t here…”). For in-domain false premises, scored 1 if the object is correctly re-targeted, and 0 if any impossible action is attempted.
  \end{itemize}

\noindent For multi-step episodes, we average the per-step Detection scores to yield a single FP score per episode.

\noindent (2) \textbf{Execution Stage.}  
Whenever IVA “Accepts,” we execute the predicted 8D joint-velocity sequence. Each RLBench task’s built-in success detector then labels the outcome as success (1) or failure (0).

\noindent (3) \textbf{Overall Accuracy.}  
We average the 225 episode scores (Detection + Execution) to yield a single accuracy metric that jointly rewards correct task execution and safe refusal/clarification behavior.

\minisection{False Premise Detection and Correction}
A major challenge in real-world robotics is handling instructions that reference unavailable objects or impossible actions. To address this, we fine-tuned {\smodel} on a curated false-premise dataset and evaluated it alongside the true-premise-trained baseline. Test prompts included both ``In-Domain'' false premises (plausible but absent objects, e.g., ``open the middle block'' when only a drawer is present) and ``Out-of-Domain'' false premises (impossible requests, e.g., ``open the top chicken'' for drawer task).

{\smodel} achieved perfect detection (100\%) on In-Domain false-premise instructions, consistently identifying and correcting these cases. For Out-of-Domain false premises, {\smodel} reached a detection accuracy of 97.78\%. In both scenarios, {\smodel} generated contextually appropriate clarifications—such as ``I don't see a tree in the current scene. Do you mean jar?''—for In-Domain false premise and terminate the interaction when face the Out-of-Domain false premise. 

\minisection{Performance on True-Premise Tasks}
To ensure that enhanced false-premise reasoning does not compromise standard performance, we tested {\smodel} on tasks with only true-premise instructions. {\smodel} maintained a success rate of $42.67\%\pm8.34\%$ compared to the baseline’s $38.67\%\pm8.55\%$. The observed difference lies well within the variance range, suggesting that the slight drop in performance is statistically insignificant. Action prediction accuracy and overall task completion remained fairly stable, confirming that robust false-premise handling does not significantly degrade general competence. 





\section{Conclusion}

In this work, we introduced the {\model} framework, enabling VLA models to robustly detect, clarify, and correct false-premise instructions in robotic settings. Our approach achieves strong performance in both false-premise detection and standard task execution, demonstrating the value of explicit false-premise reasoning in VLA models. We hope these results demonstrate the potential of language-aware robots to engage more naturally and safely with users, even when faced with ambiguous or impossible commands. Future work will focus on extending {\smodel}’s capabilities to more complex environments and real-world deployments. Finally, we do not anticipate a specific negative impact, but, as with any machine learning method, we recommend exercising caution.

\section*{Limitations}

While our IVA framework demonstrates strong performance in detecting and correcting false-premise instructions within vision-language-action (VLA) models, several limitations remain that should be addressed in future work:

\paragraph{Dataset Scope and Realism.} Our false-premise instruction dataset is primarily generated from the RLBench environment, which, despite its diversity, remains a simulated domain with a constrained set of objects, scenes, and tasks. Consequently, the distribution and complexity of false-premise instructions may not fully capture the ambiguity and diversity encountered in real-world human-robot interaction. Moreover, the proportion of In-Domain and Out-of-Domain false premises is artificially balanced to facilitate training and evaluation, and may not reflect the true incidence of such instructions in open-world settings.

\paragraph{Generalization to Real-World Deployment.} Although IVA achieves high detection rates in simulation, its robustness in real-world deployments is not yet validated. Domain shift—such as differences in visual appearance, sensor noise, or language usage—could degrade performance. Additionally, our framework assumes that visual observations and proprioceptive data are accurately and reliably captured, which may not always hold in practical robotics scenarios.

\paragraph{Correction and Clarification Strategies.} The natural language responses generated by IVA are limited to clarifications and suggestions based on the specific types of false premises represented in the training data. The model’s ability to propose truly creative or contextually appropriate alternatives remains limited, especially for Out-of-Domain or out-of-distribution instructions. In more complex environments, nuanced reasoning about task feasibility, user intent, and multi-turn clarification dialogues may be necessary.

\paragraph{Instruction and Environment Complexity.} The instructions used for evaluation are relatively short and structured, and the environments contain a modest number of distractor objects. Real human instructions can be longer, more ambiguous, and embedded in broader conversational contexts. Our current framework does not explicitly handle multi-turn dialogues, implicit user intent, or ambiguous references beyond the immediate instruction.

\paragraph{Scalability and Efficiency.} Our approach leverages instruction-tuned large multimodal models with frozen vision and language encoders, which, while effective, may impose computational and memory overhead unsuitable for some real-time or resource-constrained robotic applications.

\section*{Acknowledgments}
We would like to thank Zhang-Wei Hong and Alexander Pan for their helpful feedback and discussions.  Authors, as part of their affiliation with UC Berkeley, were supported in part by the National Science Foundation, US Department of Defense, and/or the Berkeley Artificial Intelligence Research (BAIR) industrial alliance program. This research was also developed with funding from the Defense Advanced Research Projects Agency (DARPA) under Contract No. HR0011-25-3-0133. The views, opinions and/or findings expressed are those of the author and should not be interpreted as representing the official views or policies of any sponsor, the Department of Defense, or the U.S. Government.

\bibliography{custom}

\appendix

\renewcommand{\theequation}{\thesection.\arabic{equation}}
\renewcommand{\thefigure}{\thesection.\arabic{figure}}
\renewcommand{\thetable}{\thesection.\arabic{table}}

\section*{Appendix}
\label{sec:appendix}

The appendix consists of the following further discussion:
\begin{itemize}
    \item \autoref{app:code} provides information on the code release, including links to the code bases and datasets used in the project.
    \item \autoref{app:ai_disclosure} details the use of AI in the creation of this manuscript.
    \item \autoref{app: qua_ex} visualizes examples of tasks used for training and evaluation.
    \item \autoref{app: ex_data} shows the example training dataset of In-Domain false premise, Out-of-Domain false premise, and true premise.
    \item \autoref{app: infeasible} provides the generation process of infeasible instructions.
\end{itemize}

\section{Code Release}
\label{app:code}

We make the code and data for our analysis available in our \href{https://wen-hanhsieh.github.io/iva.github.io/}{project page}. We release both under the MIT license.

\section{Disclosure of AI Usage}
\label{app:ai_disclosure}

The authors acknowledge the use of artificial intelligence (AI) tools in the preparation of this manuscript. Specifically, Microsoft Copilot, OpenAI ChatGPT, and Google Gemini Pro were utilized for general editing and code generation / completion purposes. All generated code and text was verified for correctness by one or more of the authors. 
\section{Qualitative examples}
\label{app: qua_ex}

\begin{figure*}
    \centering
    \includegraphics[width=\textwidth]{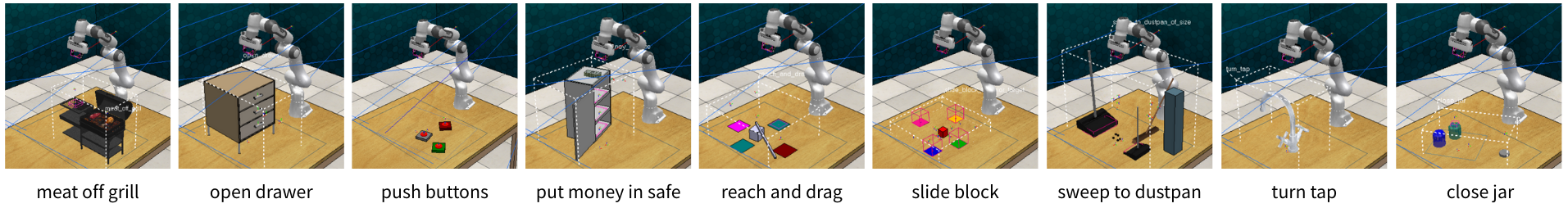}
    \caption{Qualitative examples of 9 tasks.}
    \label{fig:qualitative}
\end{figure*}

We visualize qualitative examples of 9 tasks from RLBench in \autoref{fig:qualitative}.

\section{Example Data}
\label{app: ex_data}
\begin{itemize}
    \item \textbf{In-Domain false premise:}
\begin{quote}
\small
\texttt{\textbf{human}: ``<image>\textbackslash You are a Franka robot using joint control. The task is "take the drawer off the grill", and the previous five (including current) steps are [[0, 0, 0, 0, 0, 0, 0], [0, 0, 0, 0, 0, 0, 0], [0, 0, 0, 0, 0, 0, 0], [0, 0, 0, 0, 0, 0, 0], [0.0098, 0.1741, -0.0053, -0.8438, -0.0026, 1.2311, 0.7985]]. Can you predict action of the next 1 step?''}
\end{quote}

\begin{quote}
\small
\texttt{\textbf{gpt}: ``I don't see drawer in the current scene. Do you mean chicken?''}
\end{quote}

\begin{quote}
\small
\texttt{\textbf{human}: ``Yes, You are a Franka robot using the joint control. The task is "take the chicken off the grill", and the previous five (including current) steps are [[0, 0, 0, 0, 0, 0, 0], [0, 0, 0, 0, 0, 0, 0], [0, 0, 0, 0, 0, 0, 0], [0, 0, 0, 0, 0, 0, 0], [0.0098, 0.1741, -0.0053, -0.8438, -0.0026, 1.2311, 0.7985]]. Can you predict action of the next 1 step?''}
\end{quote}

\begin{quote}
\small
\texttt{\textbf{gpt}: ``2D visual trace: [[61, 51], [61, 51], [61, 51], [61, 51], [62, 51], [60, 11], [59, 13], [57, 16], [57, 21], [55, 28], [55, 35], [56, 37], [57, 39], [59, 46], [60, 51], [58, 46], [56, 40], [57, 37], [57, 35], [61, 37], [65, 37], [72, 35], [78, 34], [80, 34], [78, 37], [76, 45], [79, 51]]. The next action step: [-0.0004, -0.0204, -0.0007, -0.0588, -0.0004, 0.0213, 0.0058, 1.0]''}
\end{quote}

    \item \textbf{Out-of-Domain false premise:}
\begin{quote}
\small
\texttt{\textbf{human}: ``<image>\textbackslash nYou are a Franka robot using joint control. The task is "take the elephant off the grill", and the previous five (including current) steps are [[0, 0, 0, 0, 0, 0, 0], [0, 0, 0, 0, 0, 0, 0], [0, 0, 0, 0, 0, 0, 0], [0, 0, 0, 0, 0, 0, 0], [-0.0152, 0.1717, -0.0032, -0.8589, 0.0182, 1.2171, 0.7744]]. Can you predict action of the next 1 step?''}
\end{quote}

\begin{quote}
\small
\texttt{\textbf{gpt}: ``I couldn't find an elephant in the current scene.''}
\end{quote}

    \item \textbf{True premise:}
\begin{quote}
\small
\texttt{\textbf{human}: ``<image>\textbackslash nYou are a Franka robot using joint control. The task is "take the chicken off the grill", and the previous five (including current) steps are [[0, 0, 0, 0, 0, 0, 0], [0, 0, 0, 0, 0, 0, 0], [0, 0, 0, 0, 0, 0, 0], [0, 0, 0, 0, 0, 0, 0], [0.0115, 0.1585, -0.0003, -0.8588, 0.0045, 1.2363, 0.8086]]. Can you predict action of the next 1 step?''}
\end{quote}

\begin{quote}
\small
\texttt{\textbf{gpt}: ``2D visual trace: [[73, 10], [78, 9], [83, 11], [90, 13], [96, 16], [100, 21], [103, 24], [106, 29], [107, 32], [109, 35], [110, 38], [109, 40], [108, 49], [108, 51], [108, 47], [110, 39], [109, 38], [100, 37], [87, 34], [71, 32], [59, 32], [57, 32], [58, 42], [57, 48]]. The next action step: [0.0173, 0.0007, -0.0033, -0.0291, -0.0006, 0.0108, -0.056, 1.0]''}
\end{quote}
\end{itemize}


\section{Infeasible Instruction Generation}
\label{app: infeasible}
To generate “infeasible” instructions, we first curate two distractor pools of nouns:
\begin{enumerate}
  \item \textbf{In-Domain:} objects that appear in our RLBench scenes (e.g., “blue safe”, “drawer”, “mug”).
  \item \textbf{Out-of-Domain:} objects never seen in the scenes (e.g., “sofa”, “durian”, “elephant”), drawn from a list generated by GPT.
\end{enumerate}
We then rewrite 85\% of the original true‐premise prompts using one of two LLM‐based rewriters:
\begin{enumerate}
  \item \emph{In-Domain FP} (65\% overall): replace the target OBJECT by sampling from the In-Domain pool, ensuring the new noun is absent from that episode’s scene.
  \item \emph{Out-of-Domain FP} (20\% overall): replace the target OBJECT with a noun sampled from the Out-of-Domain pool.
\end{enumerate}
Finally, we manually reviewed 200 randomly selected rewritten prompts to verify grammaticality and correctness.

\end{document}